\pdfoutput=1

\documentclass[11pt]{article}

\usepackage{EMNLP2022}
\usepackage{times}
\usepackage{latexsym}
\usepackage[T1]{fontenc}

\usepackage[utf8]{inputenc}
\usepackage{microtype}
\usepackage{graphicx}
\usepackage{multirow}
\usepackage{xcolor}
\usepackage{amsmath,amssymb,stackengine}
\usepackage{array}

\usepackage{svg}
\author{%
Ayal Klein%
~\;~\;~ Eran Hirsch%
~\;~\;~ Ron Eliav\\
\textbf{~\;~\;~ Valentina Pyatkin%
~\;~\;~ Avi Caciularu%
~\;~\;~ Ido Dagan}\\
  Computer Science Department, Bar-Ilan University\\%
  \texttt{\footnotesize{\{ayal.s.klein,hirsch.eran,roneliav1,valpyatkin,avi.c33\}@gmail.com}}\,\\ \texttt{\footnotesize{dagan@cs.biu.ac.il}}
}

\title{QASem Parsing: Text-to-text Modeling of QA-based Semantics}

\date{}

\begin{document}
\maketitle

\begin{abstract}
Various works suggest the appeal of incorporating explicit semantic representations when addressing challenging realistic NLP scenarios.
Common approaches offer either comprehensive linguistically-based formalisms, like AMR, or alternatively Open-IE, which provides a shallow and partial representation. 
More recently, an appealing trend introduces semi-structured natural-language structures as an intermediate meaning-capturing representation, often in the form of questions and answers.

In this work, we further promote this line of research by considering three prior QA-based semantic representations. These cover verbal, nominalized and discourse-based predications, regarded here as jointly providing a comprehensive representation of textual information --- termed \textit{QASem}.
To facilitate this perspective, we investigate how to best utilize pre-trained sequence-to-sequence language models, which seem particularly promising for generating representations that consist of natural language expressions (questions and answers). 
In particular, we examine and analyze input and output linearization strategies, as well as data augmentation and multitask learning for a scarce training data setup.
Consequently, we release the first unified QASem parsing tool, easily applicable for downstream tasks that can benefit from an explicit semi-structured account of information units in text. \end{abstract}

\section{Introduction}
\label{sec:intro}

\begin{table*}[t]
\small
\centering
\begin{tabular}{|cccc|}
\hline
\multicolumn{4}{|c|}{Both were \textbf{shot} in the \textbf{confrontation} with police and have been \textbf{recovering} in hospital since the \textbf{attack} .} \\ \hline
\multicolumn{1}{|l|}{\multirow{7}{*}{QA-SRL}} & \multicolumn{1}{r|}{1} & \multicolumn{1}{l|}{When was someone \textbf{shot?}} & in the confrontation ; the attack \\
\multicolumn{1}{|l|}{} & \multicolumn{1}{r|}{2} & \multicolumn{1}{l|}{Who was \textbf{shot?}} & Both \\
\multicolumn{1}{|l|}{} & \multicolumn{1}{r|}{3} & \multicolumn{1}{l|}{Who \textbf{shot} someone?} & police \\
\multicolumn{1}{|l|}{} & \multicolumn{1}{r|}{4} & \multicolumn{1}{l|}{Where has someone been \textbf{recovering?}} & in hospital \\
\multicolumn{1}{|l|}{} & \multicolumn{1}{r|}{5} & \multicolumn{1}{l|}{How long was someone \textbf{recovering} from something?} & since the attack \\
\multicolumn{1}{|l|}{} & \multicolumn{1}{r|}{6} & \multicolumn{1}{l|}{Who was \textbf{recovering} from something?} & Both \\
\multicolumn{1}{|l|}{} & \multicolumn{1}{r|}{7} & \multicolumn{1}{l|}{What was someone \textbf{recovering} from?} & shot \\ \hline
\multicolumn{1}{|l|}{\multirow{2}{*}{QANom}} & \multicolumn{1}{r|}{8} & \multicolumn{1}{l|}{Who \textbf{confronted} with something?} & Both \\
\multicolumn{1}{|l|}{} & \multicolumn{1}{r|}{9} & \multicolumn{1}{l|}{What did someone \textbf{confront} with?} & police \\ \hline
\multicolumn{1}{|l|}{\multirow{2}{*}{QADiscourse}} & \multicolumn{1}{r|}{10} & \multicolumn{1}{l|}{\textit{Since when} have both been \textbf{recovering} in hospital?} & since the \textbf{attack} \\
\multicolumn{1}{|l|}{} & \multicolumn{1}{r|}{11} & \multicolumn{1}{l|}{\textit{While what} were both \textbf{shot}?} & During the \textbf{confrontation} with police \\ \hline
\end{tabular}
\caption{An example sentence annotated with QASem (V1) --- QA-SRL, QANom and QADiscourse. Target predicates (verbs and nominalizations) are shown in \textbf{bold}, while QADiscourse prefixes are shown in \textit{italics}. Multiple answers are delimited by a semicolon (;). 
}
\label{tab:qasem_examples}
\end{table*}

A traditional line of research in NLP has been devoted to designing various kinds of semantic representations, that aim to explicate textual meaning with a formal, consistent annotation schema.    
Representations such as Semantic Role Labeling \cite[SRL; e.g.\ ][]{framenet_1998}, Discourse Representation Theory \cite{kamp2011DRT-discourse_repr_theory} and others \cite{copestake2005MRS, banarescu2013AMR, abend2013UCCA, oepen2015SDPsemeval, white2016UDS, bos2017groningen_meaning_bank} provide applications with an explicit account of semantic relations in a text. 
Numerous recent works illustrate how leveraging explicit representations facilitate downstream processing of challenging tasks \cite{lee2019multi, huang2021extractive, mohamed2019srl4summ, zhu2021enhancing, chen2021SRLandCoref4QA, fan-etal-2019-using}.
While traditional representations rely on pre-defined schemata or lexica of linguistic classes (e.g.\ semantic roles), 
the popular approach of Open Information Extraction \cite[OpenIE;][]{etzioni2008openIE} aims for more loosely-structured, easily attainable representations, comprised  of tuples of natural language fragments.
These light-weight structures, however, come with a cost of lacking consistency and comprehensive coverage, and do not capture deeper semantic information like semantic roles.

In a recent trend, which can be seen as an emerging mid-point between full-fledged semantic formalisms and bare-bone textual fragments, researches leverage question-answer pairs (QAs) as a representation of textual information \cite{michael2018QAMR}.
For example, several works proposed using QAs as an intermediate structure for assessing information alignment between texts, e.g.\ for evaluating summarization quality \cite{eyal-etal-2019-qa4eval_summ, gavenavicius2020qas4eval_summ, deutsch-etal-2021-qas4eval_summ} and faithfulness \cite{honovich2021qgqa4factuality_eval, durmus-etal-2020-feqa}, using a question-generation plus question-answering (QG-QA) approach.
Nevertheless, such question generation and answering models were not trained to provide a coherent representation of text meaning. 

In this work, we follow an evolving paradigm, consisting of tasks that aim to comprehensively capture certain types of predications using question-answer pairs. 
The pioneering work in this framework is Question Answer driven Semantic Role Labeling \cite[QA-SRL;][]{He2015qasrl}.
Targeting verbal predicates, QA-SRL labels each predicate-argument relation with a question-answer pair, where a natural language question represents a semantic role, while answers correspond to arguments (See Table \ref{tab:qasem_examples}).
Notably, QA-SRL was shown to subsume OpenIE, which can be derived from QA-SRL annotations by reducing them to unlabeled predicate-argument tuples \cite{stanovsky2016OIEfromQASRL}.   
This appealing QA-based framework, well suited for scalable crowdsourcing \cite{fitz2018qasrl}, has been extended to account for deverbal nominalizations \cite[QANom;][]{klein2020qanom} and for information-bearing discourse relations \cite[QADiscourse;][]{pyatkin2020qadiscourse}.
We deem these individually-presented tasks as milestones toward a broad-coverage QA-based semantic representation, which we denote as \textit{QASem}. 
To make this goal accessible, we develop a comprehensive modeling framework and release the first unified tool for parsing a sentence into a systematic set of QAs, as in Table \ref{tab:qasem_examples}. This set covers the core information units in a sentence, based on the above three predication types (verbs, nominalizations and discourse relations).\footnote{This paper presents \textsc{QASem V1}. Future versions will include QA-based tasks that capture complementary information specified by adjectival predicates and other noun modifier, which are currently at a stage of ongoing work.}

Current best models for QA-SRL/QANom and QADiscourse  \cite{fitz2018qasrl, pyatkin2020qadiscourse} are classifier-based pipelines, each targeting a specific QA format.
Predictors of relation labels (questions) use a specialized architecture that suits the task-specific question structure, and are modeled independently from relation participants (answers).   
Our work leverages recent progress in text-to-text pre-trained neural models, and specifically T5 \cite{raffel2020T5model}, for predicting QA-based annotations in a generic manner. 
Our semi-structured QASem use-case is an interesting mid-ground between natural language generation and structured prediction tasks.
A QASem output sequence includes \textit{a set} of \textit{restricted natural language} fragments (the QAs), possibly harnessing the seq2seq language generation pre-training objective rather than merely model's language understanding.


We find that fine-tuning T5 on the QA-based semantic tasks is favorable over prior approaches, producing state-of-the-art models for all the aforementioned tasks. 
Our experiments suggest that T5 is good at learning the grammar characterizing our semi-structured outputs, and that input and output linearization strategies have a significant effect on performance. 
We further explore the benefits of joint multi-task training of nominal and verbal QA-SRL. 
Our tool, including models and code, is publicly available.\footnote{We publish a unified package for jointly producing all QASem layers of annotation with an easy-to-use API --- \url{https://github.com/kleinay/QASem}. The repository also includes model training and experiments code.}

\section{Background}
\label{sec:background}

\subsection{QA-based Semantic Representation}
\label{subsec:bg-qasem}
The traditional goal of semantic representations is to reflect the meaning of texts in a formal, explicit manner \cite{abend2017sota-sem-repr}. 
SRL schemes \cite{framenet_1998, kingsbury2002propbank, schuler2005verbnet}, for example, decompose a textual clause into labeled predicate-argument relations specifying "who did what to whom", while discourse-level representations \cite{mann1987RST-rhetorical_structure_theory, kamp2011DRT-discourse_repr_theory, prasad2008PDTB2.0} capture inter-clause relations. 
Such semantic representations can be leveraged by NLP applications that require an explicit handle of textual content units for their algorithms --- for example, content selection for text generation tasks \cite{mohamed2019srl4summ, liu2015amr4abstSumm, hardy2018amr4abstractiveSumm} or information consolidation in multi-document settings \cite{liao2018AMR4MDS, pasunuru2021OpenIEgraph4MDS, chen2021SRLandCoref4QA}. 


A main drawback of these carefully-designed formalisms is their annotation cost --- since they rely on schemata of linguistically-oriented categories (e.g.\ semantic roles), dataset construction requires extensive annotator training, restricting their applicability to new text domains and new languages.

In recent years, several works proposed to remedy this annotation bottleneck by taking a more ``open-ended'' approach, capturing semantics using natural language self-explanatory terms \cite{butnariu2009NC-paraphrasing-semeval, shi2019discourse_connectives, yung2019crowdsourcing, elazar2021TNE}.
In a related trend, many recent works utilize question-answer pairs from generic QA models for soliciting a manageable, discrete account of information in a text. 
These can be used as content units for planning text generation \cite{narayan2022qa-blueprint}, or for guiding textual information alignment \cite{eyal-etal-2019-qa4eval_summ, gavenavicius2020qas4eval_summ, deutsch-etal-2021-qas4eval_summ, honovich2021qgqa4factuality_eval, durmus-etal-2020-feqa}.
In Section \ref{sec:conclusion} we discuss the limitations of such ``ad-hoc'' representations in comparison to the QA-based semantic framework which we set forth here.

This paper pursues \textit{QASem}, a systematic framework for QA-based semantic representation, based on an evolving line of research that introduced so far three concrete complementary representations --- namely, QA-SRL, QANom and QADiscourse. 
QASem can be seen as an overarching endeavor of developing a comprehensive layered representation scheme, covering all important types of information conveyed by a text. 
We now turn to present the three current building blocks of QASem. 

\subsection{QASem Tasks}
\label{subsec:bg-tasks}

\paragraph{QA-SRL}
With the goal of collecting laymen-intuitive semantic annotations, QA-SRL \cite{He2015qasrl} annotates verbs with a set of natural language QAs, where each QA corresponds to a single predicate-argument relation. 
QA-SRL questions adhere to a 7-slots template, with slots corresponding to a WH-word, the verb, auxiliaries, argument placeholders (SUBJ, OBJ1, OBJ2), and a preposition. 
The QA-SRL templates were designed to comprehensively and systematically capture all kinds of arguments and modifiers, as illustrated in Table \ref{tab:qasem_examples}  
A question is aligned with one or more answers (when a role has multiple `fillers'), each is a continuous span from the sentence. 

Beyond data collection scalability  \cite{fitz2018qasrl}, QA-SRL yields a richer argument set than linguistically-rooted formalisms like PropBank \cite{kingsbury2002propbank}, including valuable implicit arguments \cite{roit2020qasrl-gs}. 
It was also shown to subsume the popular OpenIE representation \cite{stanovsky2016OIEfromQASRL} and to enhance pre-trained encoders \cite{QuASE_HeNgRo2020}.


\paragraph{QANom}
In a follow-up work, \citet{klein2020qanom} extended the QA-SRL framework to also cover deverbal nominal predicates, which are prevalent in texts. 
First, candidate nominalizations --- nouns that have a derivationally related verb --- are extracted using lexical resources \cite{miller1995wordnet, habash2003catvar}.
QANom annotators then classify whether the candidate carries a verbal, eventive meaning in context (``The \textbf{construction} of the offices...'') or not (``...near the huge \textbf{construction}'').  
Then, predicative nominalizations undergo QA-SRL annotation, generating QAs in exactly the same format as verbal QA-SRL.
The result is a unified framework for verbs and nominalizations (See Table \ref{tab:qasem_examples}), analogous to the relationship between the PropBank \cite{kingsbury2002propbank} and NomBank \cite{meyers2004nombank} projects.

\paragraph{QADiscourse}
The relationship between propositions in a text can by itself deliver factual information. 
Several formalisms, such as Rhetorical Structure Theory \cite[RST;][]{mann1987RST-rhetorical_structure_theory} or the Penn Discourse TreeBank \cite[PDTB;][]{miltsakaki2004PDTB1.0}, have labeled inter and intra-sentential discourse relations using a taxonomy of pre-defined relation senses, e.g.\ \textsc{Contingency.Condition} or \textsc{Temporal.Asynchronous.Succession}. 
Following the QA-SRL paradigm, \citet{pyatkin2020qadiscourse} proposed to annotate discourse relations using natural language question-answer pairs (See Table \ref{tab:qasem_examples}).
They devised a list of question prefixes (e.g.\ \textit{In what case X?} or \textit{After what X?}) corresponding to a subset of PDTB relation types capturing all `informational' relations, excluding senses specifying structural or pragmatic properties of the realized passage. 
Annotators were presented with a sentence and certain heuristically extracted event targets marked in that sentence. 
They were then asked to relate such event targets with a question starting with one of the prefixes, if applicable.
The question body (after the prefix) was a copied sentence span containing one of the targets whereas the answer span contained the other. 
Different from QA-SRL and QANom, both copied spans could be slightly edited to sound grammatical and fluent.

\subsection{Relationship to Other Representations}
\label{subsec:bg-other-repr}

\paragraph{Schema-based Semantic Formalisms}

It is noteworthy that while QASem achieves a systematic coverage of semantic relations through carefully designed question templates, 
these QA-based annotations do not map directly into a formal semantic ontologies like traditional semantic representations. 
Rather, the QASem philosophy is to capture how non-professional proficient speakers perceive the semantic relations in the text and express them in a natural question-answer form.
While QASem is generally proposed as an appealing alternative to traditional (schema-based) representations, the two approaches may also be seen as complementary. 
QASem can be used in many downstream tasks that require an explicit account of semantic relation structure, which may well be represented in an ``open" natural language based form (similar to OpenIE), while including an informative signal about relation types (which OpenIE lacks). 
In other scenarios, where well-defined or fine-grained semantic distinctions are crucial, schema-based semantic formalisms like traditional SRL might be more suitable.

\paragraph{QAMR} Following a similar philosophy, \citet{michael2018QAMR} introduced Question-Answer driven Meaning Representation (QAMR), a crowdsourcing scheme for annotating  sentence semantics using QAs. 
Unlike the templated questions in QASem, QAMR consists of free-formed questions incorporating at least one content word from the sentence, along with corresponding answer spans. 
This results in a highly rich yet less controlled representation. 
Consequently, as shown by \citet[][\S 4.3]{klein2020qanom}, the QAMR annotation approach yields much less comprehensive coverage of semantic relations compared to the template-based approach of QASem.

\subsection{Prior QASem Models}
\label{subsec:bg-models}
As mentioned above, previous models for QA-SRL/QANom and QADiscourse were designed to match the specific question format of each of the tasks. 
We hereby provide further details about these models. 

Leveraging its intuitive nature, \citet{fitz2018qasrl} crowdsourced a large-scale QA-SRL dataset.
The dataset was then used for training an argument-first pipeline model for parsing the concrete QA-SRL format, comprised of a span-level binary classifier for argument detection, followed by a question generator. 
The latter is an LSTM decoder which, given a contextualized representation of the selected span, sequentially predicts fillers for the 7 slots which comprise a QA-SRL question.

Since corresponding verbs and nominalizations share the same semantic frame, but differ in their syntactic argument structure, modeling both types of predicates jointly is a non-trivial yet promising approach \cite{titov2020transferVerbtoNom}. 
Nevertheless, \citet{klein2020qanom} have only released a baseline parser, retraining the model of \citet{fitz2018qasrl} on QANom data alone. 
Their model achieves mediocre performance, presumably due to the limited amount of QANom training data, which is by an order of magnitude smaller than the training data available for verbal QA-SRL. 

\citet{pyatkin2020qadiscourse} modeled the QADiscourse task with a three-step pipeline. 
Utilizing the discrete set of question prefixes, they employ a prefix classifier, followed by a pointer generator model \cite{jia-liang-2016-pointer-generator} to complete question generation. 
Finally, they fine-tune a machine reading comprehension model for selecting an answer span from the sentence.

\paragraph{}
Differing from previous pipeline approaches, we model each of the QASem tasks using a one-pass encoder-decoder architecture.
In addition, we regard the three tasks as sub-tasks of a single unified framework, proposing a single architecture for parsing QA-based semantic annotations, also applicable for future extensions of the QASem framework.

\section{Modeling}
\label{sec:method}

\begin{table*}[t]
\small
\centering
\begin{tabular}{c|ccc|ccc|ccc}
Task & \multicolumn{3}{c|}{QA-SRL} & \multicolumn{3}{c|}{QANom} & \multicolumn{3}{c}{QADiscourse} \\
Dataset & \citeyear{fitz2018qasrl} & \multicolumn{2}{c|}{\citeyear{roit2020qasrl-gs}} & \multicolumn{3}{c|}{\cite{klein2020qanom}} & \multicolumn{3}{c}{\cite{pyatkin2020qadiscourse}} \\
Split & Train & Dev & Test &  Train & Dev & Test  & Train & Dev & Test \\ \hline
Sentences & 44476 & 1000 & 999 & 7114 & 1557 & 1517 & 7994 & 1834 & 1779 \\
Predicates & 95253 & 1000 & 999 & 9226 & 2616 & 2401 & - & - & - \\
Questions & 215427 & 2895 & 2852 & 15895 & 5577 & 4886 & 10985 & 2632 & 2996 \\
Answers & 348349 & 3546 & 3549 & 18900 & 6925 & 6064 & 10985 & 2632 & 2996
\end{tabular}
\caption{QASem Datasets Statistics. QA-SRL Training set comes from \citet{fitz2018qasrl}, while evaluation sets are from \citet{roit2020qasrl-gs}.  }
\label{tab:datasets-stats}
\end{table*}

We release a \textit{QASem tool} for parsing sentences with any subset of the QA-based semantic tasks. 
Our tool first executes sentence-level pre-processing for QA-SRL/QANom. 
It runs a part-of-speech tagger to identify verbs and nouns,\footnote{we use SpaCy 3.0 --- \url{https://spacy.io/}}
then applies candidate nominalization extraction heuristics (See \S \ref{sec:background}) followed by a binary classifier for detecting predicative nominalizations \cite{klein2020qanom}. 
Identified predicates are then passed into the QA-SRL or QANom text-to-text parsing models, while the QADiscourse model takes a raw sentence as input with no pre-processing required.
The models are described in detail in the following subsections.   


\subsection{Baseline Models}
\label{subsec:method_baselines}

We first finetune pre-trained text-to-text language models on each of the QASem tasks separately (\textsc{Baseline}). 
Unless otherwise mentioned, most modeling details specified hereafter apply also for the joint models (\S \ref{subsec:method_joint}).
We experiment both with BART \cite{lewis-etal-2020-bart} and with T5 \cite{raffel2020T5model}, but report results only for the T5 model for clarity, as we consistently observed its performance to be significantly better.
We use \texttt{T5-small} due to computational cost constraints.

Our text-to-text modeling for QA-SRL and QANom is at the \textit{predicate-level} --- given a single predicate in  context, the task is to produce the full set of question-answer pairs targeting this predicate.  
Our input sequence consists of four components --- task prefix, sentence, special markers for the target predicate, and verb-form --- as in this nominalization example:

\begin{quote}
    \textit{
    parse: Both were shot in the [PREDICATE] confrontation [PREDICATE] with police ... [SEP] confront 
}
\end{quote}

The prefix (\textit{``parse:''}) is added in order to match the T5 setup for multitask learning.
Then, the sentence is encoded together with bilateral marker tokens signaling the location of the target predicate (we report alternative methods to signal predicates in Appendix \ref{app:input_linearization}). 
At last, the verbal form of the predicate (\textit{``confront''}) is appended to the input sequence. 
This is significant for QANom, since the output verb-centered QA-SRL questions involve the verbal form of the nominal predicate. 
Verbal forms are identified during the candidate nominalization extraction phase in pre-processing, and are thus available both at train and at test time.\footnote{For verbal QA-SRL, appending the verb-form (which is the predicate itself) did not improve performance. However, in the joint verbal and nominal model, all instances are appended with a verb-form for consistency.}

Since the intended output is a \textit{set} of QAs, one can impose any arbitrary order over them.
We examine different output linearization strategies, and present our findings in Section \ref{subsec:linearization}, while the main results section (\S \ref{subsec:performance}) report the best model per dataset.
Finally, the ordered QA list is joined into a structured sequence using three types of special tokens as delimiters --- \texttt{QA}$|$\texttt{QA} separator, \texttt{Question}$|$\texttt{Answers} separator, and \texttt{Answer}$|$\texttt{Answer} separator for questions with multiple answers. 

For the QADiscourse task we train a \textit{sentence-level} model. 
The input is the raw sentence, while the output is the set of QA pairs pertaining to all targets occurring in the sentence. 
Inline with our approach in QA-SRL parsing, we prepend inputs with a new task prefix, and use special tokens as delimiters (\texttt{QA}$|$\texttt{QA}  and  \texttt{Question}$|$\texttt{Answer}). 

\subsection{Joint QASem Learning}
\label{subsec:method_joint}

Leveraging the shared output format of QA-SRL and QANom, we further train a unified model on both datasets combined (\textsc{joint}). 
Taking into account the imbalance in training set size for the two tasks, we duplicate QANom data samples by a factor of $14$, approximating a 1:1 ratio between QAs of verbal and nominal predicates (See Table \ref{tab:datasets-stats}).

It is worth mentioning that we have tested several methods for incorporating explicit signal regarding the source task (i.e.\ predicate type --- verbal or nominal) of each training instance, aiming to facilitate transfer learning. 
Our experiments include: prefix variation (e.g.\ \textit{``parse verbal/nominal:''}); typed predicate marker, i.e., having a different marker token for verbal vs.\ nominal predicates; and appending the predicate type to the \textbf{output} sequence, simulating a predicate-type classification objective in an auxiliary multitask learning framework \cite[e.g.][]{bjerva-2017-AMTL, schroder-biemann-2020-estimating-AMTL}.
Nonetheless, throughout all our experiments, uninformed joint learning of verbal and nominal predicates works significantly better.

\section{Experimental Setup}
\label{sec:experiment_setup}

\paragraph{Datasets}
We use the QADiscourse and QANom original datasets \cite{pyatkin2020qadiscourse, klein2020qanom}.
For QA-SRL, we make use of the large scale training set collected by \citet{fitz2018qasrl}.
However, prior work \cite{roit2020qasrl-gs} pointed out that their annotation protocol suffered from limited recall along with multiple, partially overlapping reference answers, hindering parser evaluation. 
For these reasons, \citet{roit2020qasrl-gs} applied a controlled crowdsourcing procedure and produced a high-quality evaluation set, dedicated for fair comparison of future QA-SRL parsers.
We adopt their annotations for validation and test.\footnote{All datasets related to the QASem paradigm have been uploaded to Huggingface's dataset hub, while unifying their data format to the extent possible --- see the datasets at \url{https://huggingface.co/biu-nlp}.}
Datasets statistics are presented in Table \ref{tab:datasets-stats}.


\paragraph{Evaluation Metrics}

For QA-SRL and QANom evaluation, we adopt the measures put forward by \citet{klein2020qanom}. The unlabeled argument detection metric (\textbf{UA}) measures how many of the predicted answers are aligned with ground truth answers, based on token overlap. 
Aligned QAs are then inspected for question equivalence to assess semantic label assignment, comprising the labeled argument detection metric (\textbf{LA}).
Consequently, \textbf{LA} figures are bounded by \textbf{UA}, as they require to match both the answer and the question to a gold QA to count as a true positive QA.
Analogously, we embrace the \textbf{UQA} and \textbf{LQA} metrics proposed by \citet{pyatkin2020qadiscourse} for QADiscourse evaluation. 
See Appendix \ref{app:metrics} for a more detailed description of the evaluation measures.

\begin{table*}[t]
\small
\centering
\begin{tabular}{cc|ccc|ccc|ccc}
\hline
 &  & \multicolumn{3}{c|}{QA-SRL Full} & \multicolumn{3}{c|}{QA-SRL Small} & \multicolumn{3}{c}{QANom} \\
 &  & P & R & F1 & P & R & F1 & P & R & F1 \\ \hline
\multirow{2}{*}{Random-Order} & \textbf{UA} & 74.1 & 58.6 & 65.5 & 65.4 & 60.0 & 62.6 & 65.0 & 52.1 & 57.9 \\
 & \textbf{LA} & 61.6 & 48.7 & 54.4 & 50.3 & 46.1 & 48.1 & 45.1 & 36.1 & 40.1 \\ \hline
\multirow{2}{*}{Role-Order} & \textbf{UA} & 76.3 & 64.4 & \textbf{69.9} & 68.4 & 59.1 & 63.4 & 61.3 & 56.8 & 58.9 \\
 & \textbf{LA} & 63.7 & 53.8 & \textbf{58.4} & 52.0 & 45.0 & 48.2 & 43.1 & 39.9 & 41.4 \\ \hline
\multirow{2}{*}{Answer-Order} & \textbf{UA} & 74.7 & 63.8 & 68.8 & 69.6 & 58.4 & \textbf{63.5} & 65.6 & 53.6 & 59.0 \\
 & \textbf{LA} & 62.5 & 53.3 & 57.6 & 53.4 & 44.9 & 48.8 & 45.7 & 37.3 & 41.1 \\ \hline
\multirow{2}{*}{All-Permutations} & \textbf{UA} & 63.1 & 64.8 & 64.0 & 66.1 & 59.1 & 62.4 & 62.7 & 53.6 & 57.8 \\
 & \textbf{LA} & 51.0 & 52.3 & 51.6 & 52.9 & 47.3 & \textbf{50.0} & 44.3 & 37.8 & 40.8 \\ \hline
\multirow{2}{*}{Fixed-Permutations} & \textbf{UA} & 75.2 & 60.0 & 66.7 & 65.8 & 58.3 & 61.8 & 62.0 & 52.8 & 57.1 \\
 & \textbf{LA} & 62.2 & 49.6 & 55.2 & 50.6 & 44.8 & 47.6 & 44.4 & 37.9 & 40.9 \\ \hline
\multirow{2}{*}{Linear-Permutations} & \textbf{UA} & 72.5 & 62.8 & 67.3 & 64.3 & 60.0 & 62.1 & 61.5 & 57.0 & \textbf{59.2} \\
 & \textbf{LA} & 60.9 & 52.7 & 56.5 & 50.5 & 47.1 & 48.8 & 43.1 & 40.0 & \textbf{41.5} \\ \hline
\end{tabular}
\caption{Output linearization experiment results for the baseline models, comparing different methods for linearizing the set of QAs into output sequence(s). \textit{QA-SRL Full} refers to training on the full QA-SRL training set, while \textit{QA-SRL Small} refers to training on a sample whose size is equivalent to QANom training set. }
\label{tab:linearization}
\end{table*}
\begin{table*}[t]
\small
\centering
\begin{tabular}{cc|ccc|ccc}
\hline
\multicolumn{1}{l}{} & \multicolumn{1}{l|}{} & \multicolumn{3}{c|}{QA-SRL Test} & \multicolumn{3}{c}{QANom Test} \\
\multicolumn{1}{l}{} &  & P & R & F1 & P & R & F1 \\ \hline
\multirow{2}{*}{Role-Order} & \textbf{UA} & 73.1 & 61.3 & 66.7 & 65.7 & 53.5 & 59.0 \\
 & \textbf{LA} & 60.5 & 50.7 & 55.2 & 49.2 & 40.1 & 44.2 \\ \hline
\multirow{2}{*}{Answer-Order} & \textbf{UA} & 76.2 & 62.4 & \textbf{68.6} & 64.9 & 54.4 & \textbf{59.2} \\
 & \textbf{LA} & 63.9 & 52.4 & \textbf{57.6} & 48.1 & 40.2 & 43.8 \\ \hline
\multirow{2}{*}{Linear-Permutations} & \textbf{UA} & 72.7 & 60.9 & 66.3 & 64.3 & 54.8 & \textbf{59.2} \\
 & \textbf{LA} & 60.7 & 50.9 & 55.4 & 48.6 & 41.4 & \textbf{44.7} \\ \hline
\end{tabular}
\caption{Output linearization experiment results for the joint QA-SRL--QANom models.}
\label{tab:linearization-joint}
\end{table*}

\paragraph{Output Set Linearization Experiment}

As stated, the output of the model is parsed into a set of question-answer pairs at post-processing. 
Thus, the ordering one applies over the linearization of QAs into an output sequence can be arbitrary. 
It is therefore appealing to examine which ordering schemes facilitate model learning more than others.\footnote{To gain a more complete perspective, we refer readers to other similar output-linearization explorations \cite{chen2021pix2seq, lopez2021simplifying}.} 
We compare a randomized order (\textbf{Random-Order}) with two consistent ordering methods.
The \textbf{Answer-Order} method orders the QAs according to answer position in the source sentence, teaching the model to ``scan'' the sentence sequentially in the search for arguments of the predicate.
Alternatively, QAs can be ordered more conceptually, with respect to the semantic role they target. 
The \textbf{Role-Order} method sorts QAs by their WH-word which is a proxy of semantic role.\footnote{We use this order: \textit{What}, \textit{Who}, \textit{When}, \textit{Where}, \textit{How}, \textit{Why}.  }

In contrast to methods that confine the model to a fixed order, one could aim to teach the model to ignore QA ordering altogether.
One way to achieve order invariance is to train over various permutations of the QA set rather than a fixed order per instance \cite{ribeiro-etal-2021-investigating}. 
In addition to order-invariance, training on multiple permutations may enhance performance from a data-augmentation perspective, especially in a realistic medium-size dataset setting.

Thus, we experiment with three permutation-based augmentation methods. 
The most straight-forward approach is to include all QA permutations of each predicate (\textbf{All-Permutations}).\footnote{To avoid memory overflow, we restrict the number of incorporated permutations by $M=10$.}
Nevertheless, in order to cope with the exponential data imbalance toward predicates with more QA pairs, an alternative method samples a fixed number of $k$ permutations for all predicates (\textbf{Fixed-Permutations}; we set $k=3$). 
On the other hand, there are reasons to assume that predicates with more QAs would be generally harder for the model to learn (see Appendix \ref{app:length-factors}). 
The third method therefore samples $n=|QAs|$ permutations for each predicate, producing linearly imbalanced training data in which instance frequency is proportional to the number of QAs in its output (\textbf{Linear-Permutations}).

We train QA-SRL and QANom baseline models using each of the above mentioned linearization methods.
These models differ both in the semantic task they tackle (i.e.\ verbs vs.\ nominalizations) and in the training data scale; thus, in order to distinguish these two effects, we also experiment with training on a random subset of the verbal QA-SRL training set with the same size as the QANom training set (\textbf{QA-SRL small}). 
Results of comparing the different linearization methods are in Section \ref{subsec:linearization}.

\paragraph{Training Details}
We tuned the models' hyper-parameters on the validation sets with a grid search, detailed in Appendix \ref{app:out_linearization_details}.
The joint QA-SRL and QANom models were tuned to optimize QANom validation measures.

\section{Results}
\label{sec:results}

In this section, we present the experiments we conducted on the QASem tasks and the corresponding results. 
We start with results of the experiment testing different linearization methods, and then discuss final performance of best models. 
We conclude by assessing out-of-domain generalization.

\begin{table*}[t]
\small
\centering
\begin{tabular}{cc|ccc|ccc}
\multicolumn{1}{l}{} & \multicolumn{1}{l|}{} & \multicolumn{3}{c|}{QA-SRL} & \multicolumn{3}{c}{QANom} \\
model & \multicolumn{1}{l|}{} & P & R & F1 & P & R & F1 \\ \hline
\multirow{2}{*}{\citet{fitz2018qasrl}} & \textbf{UA} & 79.1 & 60.1 & 68.3 & 45.1 & 61.5 & 52.0 \\
 & \textbf{LA} & 53.8 & 40.9 & 46.4 & 29.6 & 40.4 & 34.2 \\ \hline
\multirow{2}{*}{T5 baseline} & \textbf{UA} & 76.3 & 64.4 & \textbf{69.9} & 61.3 & 57.5 & \textbf{59.4} \\
 & \textbf{LA} & 63.7 & 53.8 & \textbf{58.4} & 44.6 & 41.8 & 43.1 \\
\multirow{2}{*}{T5 joint} & \textbf{UA} & 76.2 & 62.4 & 68.6 & 64.3 & 54.8 & 59.2 \\
 & \textbf{LA} & 63.9 & 52.4 & 57.6 & 48.6 & 41.4 & \textbf{44.7}
\end{tabular}
\caption{Final results of parsing verbal QA-SRL and nominal QA-SRL (QANom). Test sets are from \cite{roit2020qasrl-gs} and \cite{klein2020qanom} respectively. }
\label{tab:results-qasrl}
\end{table*}
\begin{table*}[t]
\small
\centering

\begin{tabular}{c|ccc|cc}
\hline
 & \multicolumn{3}{c|}{UQA} & \multirow{2}{*}{\begin{tabular}[c]{@{}c@{}}LQA \\ Accuracy\end{tabular}} & \multirow{2}{*}{\begin{tabular}[c]{@{}c@{}}Prefix \\ Accuracy\end{tabular}} \\
 & P & R & F1 &  &  \\ \hline
\citet{pyatkin2020qadiscourse} & 80.8 & 86.8 & 83.7 & 66.6 & 49.9 \\
Ours (T5) & 87.0 & 84.3 & \textbf{85.6} & \textbf{73.3} & \textbf{57.8} \\ \hline
\end{tabular}
\caption{Evaluation results on the QADiscourse test set. }
\label{tab:results_qadiscourse}
\end{table*}

\subsection{Linearization Experiment} 
\label{subsec:linearization}

As can be seen in Table \ref{tab:linearization},
selecting a coherent ordering scheme (\textbf{Role-Order} or \textbf{Answer-Order}) consistently improves performance over the random-order baseline. 
In addition, augmenting the training data with permutations, especially using a linear bias toward longer sequences, enhances performance for QANom, but is harmful for QA-SRL.\footnote{We have also applied the permutation-based methods on QADiscourse; however, none of these improved performance over the baseline model.} 
This may be attributed to some extent to the difference in train set scale --- when abundant training samples are available, data augmentation is less effective and has lower priority compared to output's structural consistency. 
However, the ``medial'' effect on \textbf{QA-SRL small}, where augmentation methods exhibit moderate deterioration, suggest that the contrast might also be attributed to the verbal vs.\ nominal distinction; for example, to nominalizations' more flexible argument structure \cite{alexiadou2010nominalizations}, positing output order consistency less effective than for verbal predicates. 

The latter conjecture is supported by an additional linearization experiment we applied on the joint learning setting, whose results are shown in Table \ref{tab:linearization-joint}. 
While testing on nominal predicates favors the order-invariant, permutation-based method, the same model benefits the most from the \textbf{Answer-Order} method when testing on verbal predicates.     

Overall, our experiment indicates that linearization techniques have a substantial effect on predicting semi-structured outputs (e.g.\ sets) with seq2seq models. 
In the next subsection, we compare our best models to prior QASem models.

\subsection{Models Performance}
\label{subsec:performance}

\paragraph{QA-SRL and QANom}
Table \ref{tab:results-qasrl} presents evaluation measures of the best performing model per setting from the previous subsection.\footnote{
That is --- \textbf{Linear-Permutations} for the QANom models, \textbf{Role-Order} for QA-SRL Baseline, and \textbf{Answer-Order} for the joint model tested on QA-SRL.}
We can see that the T5-based models are improving over the previous approach with a noticeable margin, especially with respect to question quality (\textbf{LA}).  
Notably, the argument-detection (\textbf{UA}) improvement for QANom is much more profound than for QA-SRL. 
We ascribe this to its smaller training size, putting more weight on the pre-training phase.\footnote{The model version we used for the prior QA-SRL model \cite{fitz2018qasrl} is using ELMo contextualized embeddings \cite{peters2018ELMo}, which although belonging to the pre-trained language-model regime, are significantly weaker compared to more recent PLMs \cite{devlin2019bert}.}  

As for the joint learning of verbal and nominal predicates, it seems to have a positive effect only for question quality in the nominal domain. 
This can also be attributed to training size --- whereas verbal QA-SRL is slightly impaired from adding nominal instances to the training data, the benefit of nominal predicates from significantly enlarging the training set overcomes this adverse effect.

Overall, turning to T5 improved both QA-SRL and QANom \textbf{LA} F1 performance by over 25\% compared to previous state-of-the-art parsers, while joint learning gains another 9\% recall and 4\% F1 for QANom.\footnote{Taking memory efficiency into account, our QASem tool uses the \textbf{Answer-Order} joint model for both QA-SRL and QANom by default, fetching a single model for both types of predicates.}

\paragraph{QADiscourse}

Performance evaluation of our QADiscourse model over the QADiscourse task, compared to the previous pipeline model \cite{pyatkin2020qadiscourse}, is reported in Table \ref{tab:results_qadiscourse}.
While unlabeled detection of discourse relations is improving by a relatively small margin, the question quality --- assessed by the LQA and prefix accuracy metrics --- is substantially increased. 
Results suggest that the model is leveraging the generative language modeling pre-training, possibly making its generated question-answer statements more semantically sound, as may also be entailed from the large increase in precision (8\%).

\subsection{Out-of-Domain Generalization}
\label{subsec:ood}

Finally, to estimate the expected performance of our parser in a realistic downstream scenario, we conducted an experiment tackling out-of-domain generalization.
The QA-SRL training set, taken from the large-scale QA-SRL corpus released by \citet{fitz2018qasrl}, includes 3 considerably diverse domains --- encyclopedic (\textsc{Wikipedia}), news (\textsc{Wikinews}) and scientific text books (\textsc{TQA}). 
The evaluation set is comprised only of the first two. 
While the models reported so far were trained on all available domains, in order to compare in-domain and out-of-domain generalization more carefully, we trained models on each domain separately and evaluated against single-domain test sets. 
We bound the training set sizes to that of \textbf{QA-SRL Small} (for comparability with Table \ref{tab:linearization}), use \textbf{Answer-Order} linearization, and perform the same grid hyper-parameter search procedure (Appendix \ref{app:out_linearization_details}).

Results (Table \ref{tab:cross-domain}) indicate that while best performance is obtained using in-domain training data, out-of-domain performance decreases by merely 3.0--0.7 F1 points. This implies that models generalize quite robustly to out-of-domain corpora, which is encouraging for downstream usage.

\section{Analyses}
\label{sec:analyses}

\paragraph{Output Validity} 
As mentioned in Section \ref{subsec:bg-tasks}, QA-SRL questions adhere to a specialized  constrained format. 
It is therefore not trivial for a model pre-trained on free natural language to acquire these format specifications.
Nevertheless, we observe that the models have robustly internalized the special grammar of the QA sequences. 
Only a small fraction ($1.2\%$) of output QA-SRL/QANom QAs were automatically detected as not conforming with QA-SRL specifications, of which vast majority ($>95\%$) are due to answer--sentence mis-alignment mostly owing to tokenization issues (e.g.\ answer token is out-of-vocabulary).


\paragraph{Manual Error Analysis}
Prior works on QA-SRL have acknowledged that the automatic evaluation metrics are under-estimating true performance figures \cite{roit2020qasrl-gs, klein2020qanom}.
We inspected the joint model predictions on the verbal and nominal QA-SRL test sets, taking samples of 50 QAs automatically classified as precision mistakes, and of 50 gold-standard QAs classified as recall misses (200 QAs total).

Our findings are detailed in Appendix \ref{app:manual-error-analysis}.
To summarize the verbal QA-SRL findings, we conclude that 42\% of the precision mistakes are actually acceptable answers, whereas 40\% of counted recall mistakes have correct counterparts in model predictions, both of which erroneously rejected by the strict alignment-based evaluation metric. 
Acceptable mistakes are often caused by taking  different span-selection decisions, or by an argument structure having multiple correct interpretations.
Unacceptable mistakes commonly concern answer-repetition, verb-particle constructions, and missing harder implied arguments.   
Overall, considering this manual analysis, the joint model interpolated \textbf{UA} precision on QA-SRL is \textbf{87.0} while recall is \textbf{78.6}. 
Interpolated \textbf{UA} for QANom is much lower --- \textbf{77.2} precision, \textbf{62.0} recall --- leaving room for future improvements.  
 
\begin{table}[t]
\small
\centering
\begin{tabular}{ccc|ccc}
Train Domain                        & Test Domain                         & \multicolumn{1}{l|}{} & P             & R             & F1            \\ \hline
\multirow{2}{*}{\textit{Wikipedia}} & \multirow{2}{*}{\textit{Wikipedia}} & UA                    & \textit{71.0} & \textit{57.8} & \textit{63.7} \\
                                    &                                     & LA                    & \textit{58.0} & \textit{47.2} & \textit{52.1} \\
\multirow{2}{*}{TQA}                & \multirow{2}{*}{Wikipedia}          & UA                    & 72.2          & 55.7          & 62.9          \\
                                    &                                     & LA                    & 58.4          & 45.1          & 50.9          \\
\multirow{2}{*}{Wikinews}           & \multirow{2}{*}{Wikipedia}          & UA                    & 72.0          & 56.0          & 63.0          \\
                                    &                                     & LA                    & 58.3          & 45.3          & 51.0          \\ \hline
\multirow{2}{*}{\textit{Wikinews}}  & \multirow{2}{*}{\textit{Wikinews}}  & UA                    & \textit{74.4} & \textit{66.3} & \textit{70.1} \\
                                    &                                     & LA                    & \textit{61.1} & \textit{54.4} & \textit{57.5} \\
\multirow{2}{*}{TQA}                & \multirow{2}{*}{Wikinews}           & UA                    & 73.4          & 63.8          & 68.3          \\
                                    &                                     & LA                    & 58.6          & 51.0          & 54.5          \\
\multirow{2}{*}{Wikipedia}          & \multirow{2}{*}{Wikinews}           & UA                    & 68.4          & 65.9          & 67.1          \\
                                    &                                     & LA                    & 55.4          & 53.3          & 54.3         
\end{tabular}
\caption{QA-SRL evaluation results of in-domain (\textit{italics}) vs.\ out-of-domain test settings.}
\label{tab:cross-domain}
\end{table}
 
 \paragraph{QA Position Effect} 
 A further analysis, reported in Appendix \ref{app:length-factors}, examines the effect of position in generated sequence on QA quality. 

\section{Conclusion}
\label{sec:conclusion}

We propose to bundle three QA-based semantic tasks into a congruent conceptual paradigm.
We hence develop and release new state-of-the-art models for these tasks, based on a unified framework for fine-tuning a seq2seq pre-trained language model. 
Specifically, we show the importance of output linearization choices, including permutation-based data augmentation techniques, and propose using joint learning of verbal and nominal QA-SRL for further enhancing performance in medium-size dataset settings.
We further demonstrate these models' out-of-domain robustness. 

Utilizing these models, the QASem tool we release can be used in various downstream scenarios where an explicit account of textual information units is desired. 
For example, the recent trend of leveraging QAs as an intermediate representation for various summarization-related tasks indicates the perceived attractiveness of this ``open" representation style. 
In particular, questions and answers provide a natural linguistic mechanism for explicitly focusing on concrete information units. 
However, the common QA datasets \cite[e.g.\ SQuAD;][]{rajpurkar-2016-squad}, over which prior QA representations have been trained, were developed for modeling QA or reading comprehension as end-tasks, but were not designed to provide a systematic semantic representation. 
Hence, QA models trained on such datasets yield a non-systematic set of QAs, which might introduce overlapping and non-exhaustive information units, hindering their downstream utility.
On the other hand, QASem is designed to produce a systematic --- i.e.\ consistent and comprehensive --- set of QAs, 
each targeting an atomic ``statement" concerning different predications, thus 
providing a more precise representation of semantic structure.

Future work would incorporate upcoming QASem tasks regarding adjectives and noun modifiers into the current seq2seq framework. 
Further, we plan to explore sentence-level modeling for predicting all QASem QAs jointly.

\section{Limitations}
\label{sec:limitations}

Our QASem model is built upon vanilla T5. 
Nevertheless, many question-answering datasets exist, which could quite probably enhance QASem parsing through some multitask or transfer learning setting. 
Although we have had a preliminary experiment with a pre-trained question-generator model, yielding negative results, a more careful exploration of this path seems promising. 
In particular, one could leverage QA datatsets to pre-train a text-to-text model on QA-generation in the same output format as our QASem tasks. 

An essential limitation of the current approach is that model outputs do not assign any confidence score for generated QAs. 
This seem like a crucial feature to have for deployment in downstream systems, e.g.\ for controlling over the precision/recall trade-off. 
As posterior probabilities of generated tokens are conditioned on all previous tokens in the sequence, it is not trivial to deduce a confidence score for a sub-sequence. 
Hence, the challenge of confidence estimation for semi-structured predictions using seq2seq  warrants further research. 

\section*{Acknowledgements}
This work was supported in part by grants from Intel Labs, the Israel Science Foundation grant 2827/21, and the PBC fellowship for outstanding PhD candidates in data science.

\bibliography{00_main_body.bib}
\bibliographystyle{acl_natbib}

\appendix
\section{Appendices}
\label{sec:appendix}

\subsection{Detailed Evaluation Metrics}
\label{app:metrics}

Evaluating QA-based semantic tasks involves two core aspects. 
First, we would like to estimate how many of the \textit{semantic relations} are captured correctly.
For SRL, this is analogous to measuring argument detection, while for discourse, it assesses whether pairs of events are related to each other or not.
Second, given that the model identified the same predicate-argument or predicate-predicate relation as present in the gold set, we want to assess its predicted label for the relation type (semantic role or discourse relation sense).
A manifestation of these objectives for the QA-SRL and QADiscourse formats considers an \textit{unlabeled} and a \textit{labeled} evaluation measure per task \cite{roit2020qasrl-gs, pyatkin2020qadiscourse}. 

For computing QA-SRL's unlabeled argument detection (\textbf{UA}) metric, QAs in the predicted set are aligned to QAs in the reference set using maximum bipartite matching based on lexical intersection-over-union (IOU) of the answers. 
A pair of QAs must surpass a minimum IOU threshold $\Gamma$ to count as aligned.
Then, aligned QA pairs are re-inspected for question equivalence to form the labeled argument detection measure (\textbf{LA}). 

QA-SRL question templates have no plain mapping to semantic roles, and determining whether two questions refer to the same role is non-trivial.  
Thus, previous QA-SRL works have proposed different heuristics for evaluating approximated question equivalence. 
Here we apply the evaluation measures put forward by \citet{klein2020qanom}, using a technique for mapping questions into a discrete space of ``syntactic roles'', and setting $\Gamma=0.3$. 
We apply it on both QA-SRL and QANom to have comparable figures. 

As for QADiscourse, we simply embrace the \textbf{UQA} and \textbf{LQA} metrics proposed by \citet{pyatkin2020qadiscourse}. 
These are analogous to \textbf{UA} and \textbf{LA}, with minor adaptations. 
The unlabeled alignment between QA pairs is computed as IOU between question-and-answer tokens jointly ($\Gamma=0.5$), excluding question prefix, because the question words denote which proposition is participating in the discourse relation with the answer.
In addition, labeled alignment is simply a match over question prefixes, since unlike QA-SRL question, these question prefixes do map into relation senses.

\subsection{Alternative QA-SRL Input Linearization Methods}
\label{app:input_linearization}
Here we specify in greater detail about experiments we ran assessing alternative linearization methods for QA-SRL and QANom models. 

Concerning the input encoding, we experimented with four methods of highlighting the target predicate token within the sentence:
\begin{enumerate}
    \item Repeating the target word at the end of the sequence
    \item Special token before the target 
    \item Special token after the target
    \item Special tokens before and after the target
\end{enumerate}
Method 4.\ outperformed methods 2.\ and 3.\ by a small margin, while method 1.\ was worse.

\subsection{Training Details}
\label{app:out_linearization_details}

In our preliminary experiments, model training was shown to be quite sensitive to hyper-parameter tuning.  
Nevertheless, it is impractical to execute a wide hyper-parameter search to test each linearization method. 
Instead, for the small training-set experiments (QANom and \textbf{QA-SRL Small}) we constrained the tuning phase to a small grid search:
$$ \text{learning rate} \in \{0.001, 0.005, 0.01\} $$
$$ \text{dropout rate} \in \{0.1, 0.15\} $$ 
$$ \text{effective batch size} \in \{96, 168\} $$ 
As the training set of \textbf{QA-SRL Full} is 14-times larger, even this grid-based method has been unaffordably expensive. This also applies for  the joint model's training process. Thus, for these settings we fix the hyper-parameters throughout all linearization methods, using:
$$ \text{learning rate} = 0.005 $$
$$ \text{dropout rate} = 0.1  $$
$$ \text{effective batch size} = 96 $$ 
All models were fine-tuned for 20 epochs, with \texttt{fp16} mode, and used a beam size of 5 for decoding. 

\begin{figure*}[t]
    \centering
    \includegraphics[width=145mm]{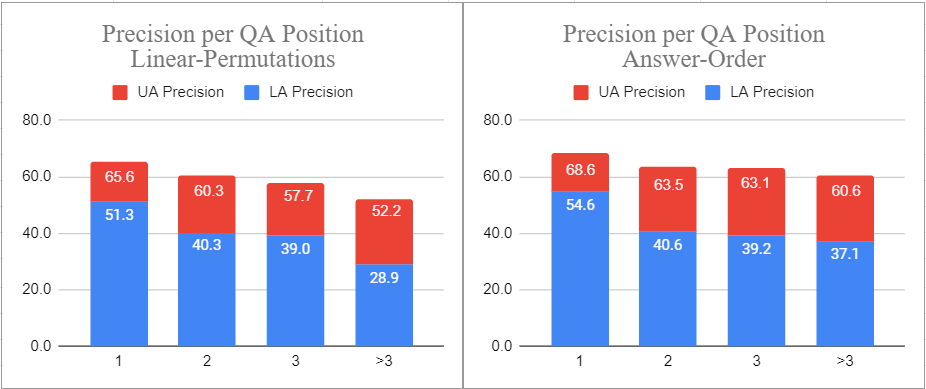}
    \captionof{figure}{Predicted QA Precision (y axis) per QA position in output sequence (x axis).}
    \label{fig:qa-position}
\end{figure*}

\subsection{Manual Error Analysis}
\label{app:manual-error-analysis}
As mentioned in Section \ref{sec:analyses}, we have manually inspected the joint model predictions on the both (verbal) QA-SRL and nominal QA-SRL (QANom) test sets. For each task, we took a sample of 50 QAs automatically classified by the UA measure as precision mistakes, and a sample of 50 gold-standard QAs classified as recall misses.

\paragraph{QA-SRL}
We judged 21 / 50 of precision mistakes (42\%) as  acceptable answers, and 20 / 50 (40\%) of recall mistakes as having correct counterparts in model predictions.

These are mostly characterized by the fact that the model concatenates answers while the gold-standard has a better separation of answers. For example, the gold-standard  contains the pair \textit{Q: Who pleaded something?  A: ['Co-defendant', 'Daniel Spitler']}, while the model's prediction has the same question with the concatenated version of the answer \textit{'Co-defendant Daniel Spitler'}. 
Another common type of the acceptable mistakes is where two QAs (i.e.\ roles) can be alternatively captured by a single QA. 
For example, for the sentence \textit{The company also \textbf{announced} Daniel Ammann as its new president}, the gold-standard contains: \textit{Q: Who did someone announce as something?  A: Daniel Ammann  ;  Q: What did someone announce someone as?  A: its new president}. In contrast, the model predicts \textit{Q: What did someone announce?  A: Daniel Ammann as its new president}.


With respect to genuine mistakes, some precision errors occur in sentences with phrasal verbs, such as \textit{'come across'} or \textit{'carry out'}, where the model fails to ask the correct question using the verb particle construction. 
On the other hand, we observed that several recall errors are regarding adjuncts occurring in a non-standard position; for instance, for the sentence: \textit{The top deck of the bus was crushed on one side after hitting the truck and \textbf{spinning}}, the models misses the following gold QA --- \textit{Q: When did something spin?  A: after hitting the truck}. 
Quantitatively, gold-standard questions starting with \textit{Why} or \textit{How} have a better chance of being missed by the model, in line with their stronger reliance on common-sense reasoning skills.    

\paragraph{QANom}

The automatic evaluation for QANom have been more accurate. 
We judged 18 / 50 of precision errors (36\%) as acceptable QAs, and only 8 / 50 of recall errors (16\%) as having correct counterparts in model predictions.

For QANom, acceptable precision mistakes are often due to incomplete coverage of the gold annotations. 
For example, annotations for the sentence \textit{Alex Neil, the Scottish cabinet \textbf{minister} responsible for the legislation, said: `` This is a historic moment for equality in Scotland''} are missing the following model-generated QA --- \textit{Q: Where did someone minister something?  A: Scotland}. 
Another common cause of acceptable mistakes are slight variations in phrasing in the question-answer pair. 
An example is the following gold-standard QA --- \textit{Q: Where did someone legislate?  A: In Scotland} --- compared to the following prediction: \textit{Q: Where did someone legislate something?  A: Scotland}.

The genuine precision mistakes are  characterized by the model generating questions that have no answer in the sentence, thus aligning it to an unfaithful answer. 
For example, for the predicate \textit{Prosecutors claim political \textbf{assassinations} and suicide attacks were planned}, one of the model-generated QAs is \textit{Q: Who assassinated someone?  A: Prosecutors}. Once such a question is generated, a generation of an answer will inevitably lead to a mistake. 
Similarly to QA-SRL, recall mistakes commonly concern implicit arguments, which are more frequent at the QANom dataset compared to QA-SRL \cite{klein2020qanom}. For example, for the sentence \textit{As a \textbf{protest} against the punishment, Issawi began a publicized hunger strike}, the model misses the following gold-standard QA --- \textit{Q: How did someone protest?  A: began a publicized hunger strike}.

\subsection{QA-Position Impacting Model Precision}
\label{app:length-factors}

In this section we investigate how QA position in output sequence affects generation quality, and whether output linearization methods interact with these effects.

Taking QANom-Baseline as our model, we analyze the precision of predicted QAs with respect to their position in the output sequence. 
Results for the \textbf{Answer-Order} and \textbf{Linear-Permutations} output linearization methods are plotted in Figure \ref{fig:qa-position}. 
There is a clear effect of the QA's position on its accuracy --- QAs generated first by the auto-regressive decoder have higher quality than those generated last.
A consequence, also quantitatively observed in model predictions, is that predictions for predicates with many true arguments would have lower precision than those with few arguments.

Interestingly, the above mentioned effect is mitigated when training on a fixed linearization order (\textbf{Answer-Order}) rather than on permutations. 
This may be caused by the fact that, following the fixed order of QAs with respect to answer position in sentence seen during training, the model is learning to ``constrain'' itself to predicting spans from the ``remaining'' part of the sentence, narrowing its false-positive choices.     


\end{document}